\documentclass[journal,onecolumn,draftcls]{IEEEtran}

\newcommand{\StyleDraft}[0]{\StyleFinalfalse}
\newif\ifStyleFinal\StyleFinaltrue
\StyleDraft

\usepackage{calc}
\ifStyleFinal
	\newcommand{\figwidth}{\minof{\columnwidth}{5.0in}}
	\usepackage{subcaption}
\else
	\newcommand{\figwidth}{\minof{\columnwidth}{6.0in}}
	\usepackage{subfig}
\fi

\usepackage{cite}
\usepackage{graphicx}
\graphicspath{{./figs/}}

\ifCLASSINFOpdf
  \usepackage{graphicx}
  \graphicspath{{./figs/}}
\else
\fi

\usepackage[cmex10]{amsmath}
\usepackage{multirow}
\usepackage{booktabs}

\usepackage{breqn}

\usepackage{caption}
\usepackage{tabularx}
\begin{document}
\title{Real-Time Impulse Noise Suppression from Images Using an Efficient Weighted-Average Filtering}

\author{Hossein~Hosseini, Farzad Hessar,~\IEEEmembership{Student Member,~IEEE} and Farokh Marvasti,~\IEEEmembership{Senior Member,~IEEE}
\thanks{Hossein~Hosseini and Farokh Marvasti are with the department of Electrical Engineering, Sharif University of Technology. email: h\_hosseini@alum.sharif.edu, marvasti@sharif.edu}%
\thanks{Farzad Hessar is with the department of Electrical Engineering, University of Washington, USA. email: farzad@u.washington.edu}
}

\maketitle
\thispagestyle{empty}

\begin{abstract}
In this paper, we propose a method for real-time high density impulse noise suppression from images. In our method, we first apply an impulse detector to identify the corrupted pixels and then employ an innovative weighted-average filter to restore them. The filter takes the nearest neighboring interpolated image as the initial image and computes the weights according to the relative positions of the corrupted and uncorrupted pixels. Experimental results show that the proposed method outperforms the best existing methods in both PSNR measure and visual quality and is quite suitable for real-time applications.
\end{abstract}


\IEEEpeerreviewmaketitle

\section{Introduction}
Images are often corrupted by impulse noise during acquisition and transmission. Therefore, an efficient noise suppression method is required before subsequent image processing operations \cite{book:Bovic:2000}. 
Many recent methods \cite{paper:Chan:2005, paper:Srinivasan:2007, paper:Deng:2007, paper:Chen:2008, paper:Yildirim:2008, paper:Zhang:2009, paper:Toh:2010, paper:Fabijanska:2011, paper:Esakkirajan:2011, paper:Xiong:2012, paper:Zhou:2012, paper:Chou:2013, paper:Hosseini:2013} first detect the corrupted pixels and then restore them without affecting the uncorrupted pixels. Various  solutions for estimating the intensity of noisy pixels can be divided into four categories of median-based filters, fuzzy-based algorithms, adhoc ideas and weighted-average filters. 

Most of impulse noise removal algorithms are variations of median filtering. Best examples are Decision-Based Algorithm (DBA) \cite{paper:Srinivasan:2007}, Median-based Switching filter (MS) \cite{paper:Fabijanska:2011}, and Modified Decision Based Unsymmetric Trimmed Median Filter (MDBUTMF) \cite{paper:Esakkirajan:2011}. Also, due to the nature of impulse noise, some methods are proposed based on fuzzy logics, such as Detail-Preserving Filter (DPF) \cite{paper:Yildirim:2008}, Noise Adaptive Fuzzy Switching Median (NAFSM) filter \cite{paper:Toh:2010}, and Turbulent Particle swarm optimization based Fuzzy Filtering (TPFF) \cite{paper:Chou:2013}.

Other methods employed different ideas. In \cite{paper:Chan:2005}, Specialized Regularization (SR) method is proposed to restore noisy pixels. Opening-Closing Sequence (OCS) filter is presented in \cite{paper:Deng:2007} based on mathematical morphology. In \cite{paper:Chen:2008}, Edge-Preserving Algorithm (EPA) is proposed which adopts a directional correlation-dependent filtering technique. In \cite{paper:Xiong:2012}, Robust Outlyingness Ratio is combined with the Non-Local Means (ROR-NLM) to detect and filter the noisy pixels. 
In \cite{paper:Hosseini:2013}, a method is presented which employs an iterative impulse detector and an Adaptive Iterative Mean (AIM) filter to remove the general fixed-valued impulse noise.

Another well-known approach is weighted-average filtering, which exploits the correlation among neighboring pixels to restore the corrupted pixels. The Switching-based Adaptive Weighted Mean (SAWM) filter \cite{paper:Zhang:2009} and the Cloud Model (CM) filter \cite{paper:Zhou:2012} employ this approach for impulse suppression. Both filters adaptively determine the filtering window and use complex weighting rules. In SAWM method, the weights are specified based on the degree of compatibility between pixels, and the CM filter uses the certainty degrees of uncorrupted pixels as their weights.
These filters are time-varying; that is they have to perform pixel-by-pixel restoration, rather than processing the image as a whole. This constraint opposes efficient implementation.

In this paper, we propose a two-step method for real-time impulse noise suppression. First, we employ an impulse detector to identify the corrupted pixels. It examines the spatial correlation of suspicious image pixels to decrease the false detection of uncorrupted pixels as corrupted. Second, we restore the image using a weighted-average filter. The filter operates on the nearest neighboring interpolated image and can be implemented using matrix-based operations. 

The rest of this paper is organized as follows. Section II defines the impulse noise model. The method is presented in section III. The experimental results and comparisons are provided in section IV and section V concludes the paper.

\section{Impulse Noise Model}
Fixed-Valued Impulse Noise (FVIN), also known as Salt-and-Pepper Noise (SPN), is commonly modeled by:
\begin{align}
\tilde{x}_{i,j}=\left\{ 
\begin{array}{lcr}
N_{\min} & 	\mbox{with probability} & p/2 \\
x_{i,j}&  \mbox{with probability} & 1-p \\
N_{\max} &  \mbox{with probability} & p/2
\end{array}
\right.
\end{align}
where $x$, $\tilde{x}$ and $p$ are the original and corrupted images and noise density, respectively, and $(i,j)$ is the image coordinate. This model implies that the pixels are randomly corrupted by two fixed extreme grey-values, $N_{\min}$ and $N_{\max}$, with the same probability.

For impulse noise suppression, we first specify the impulse values and locate the corrupted pixels, and then estimate their original values using the information provided by the uncorrupted pixels.

\section{The Proposed Method \cite{code:Hossein:2014}}
The proposed noise suppression method consists of two steps: Impulse Detection and Image Restoration. Each step is further discussed in the following.

\subsection{Impulse Detector}
To identify the corrupted pixels, we first determine the impulse values, $N_{\min}$ and $N_{\max}$. However, marking all pixels with an extreme grey-value as noisy pixels results in a false detection of some uncorrupted pixels as corrupted. Therefore at the next step, we should locate the noisy pixels by discriminating the uncorrupted pixels which have an impulse value. For this, we use the fact that SPN alters the pixel values to one of the two extreme grey-values with equal probabilities. Thus, a strong inclination toward one of the impulse values in a neighborhood indicates that there are some uncorrupted pixels with an impulse value. We examine the inclination for each neighborhood and the correlation of each suspicious pixel with its neighbors to distinguish between the corrupted pixels and the uncorrupted ones which have one of the impulse values.

\textbf{Computing the window size:} Given the estimated noise probability is $\tilde{p}$, for a pixel the probability of being uncorrupted is approximately $1-\tilde{p}$. Thus, using binomial distribution, the expected number of uncorrupted pixels in a window of size $w\times w$ is $(w^2-1)(1-\tilde{p})$. If we set this value equal to five, the window size is obtained as the smallest odd integer greater than $\sqrt{1+\frac{5}{1-\tilde{p}}}$. 
The window size is specified such that the expected number of uncorrupted pixels in the pixels' neighborhood becomes five, and, as a result, there would be enough uncorrupted pixels to examine the center pixel.

We describe the details of the impulse detector in the following: 
\begin{enumerate}
	\item Specify the impulse values, $N_{\min}$ and $N_{\max}$, by finding the two extreme grey-values of the corrupted image, and then construct the set $\Omega_N$ as: 
	\begin{align}
		\Omega_N=\left\{ (i,j)  \left| (\tilde{x}_{i,j}=N_{\min}) \vee (\tilde{x}_{i,j}=N_{\max}) \right. \right\}
	\end{align}
where the symbol $\vee$ denotes the logical OR. The set $\Omega_N$ includes the indices of suspicious pixels, i.e. pixels with one of the impulse values. 

	\item Compute the estimated noise probability $\tilde{p}$ as the rate of the suspicious pixels and set $w$ equal to the smallest odd integer greater than $\sqrt{1+\frac{5}{1-\tilde{p}}}$. 
	
	\item Compute $c_{i,j}^\min$ and $c_{i,j}^{\max}$ as the number of pixels, in the neighborhood of the pixel at coordinate $(i,j)$, with grey-values equal to $N_{\min}$ and $N_{\max}$, respectively.

	\item Construct the sets $\Omega_{d_1}$ and $\Omega_{d_2}$ as follows: 
\ifStyleFinal
	\begin{align}	
	\label{eq:OmegaD}
		\Omega_{d_1} =& \left\{ (i,j) \left| \vphantom{c_{i,j}^{\min}} \right.\right. \left. c_{i,j}^{\min}+c_{i,j}^{\max}=w^2 \right\}\\
		\Omega_{d_2} =& \left\{ (i,j) \left| \left((\tilde{x}_{i,j}=N_{\min})\wedge(c_{i,j}^{\max}<\frac{c_{i,j}^{\min}}{3})\right) \vee  \right.\right. \nonumber \\
& \ \  \left.  \vphantom{\frac{c_{i,j}^{\min}}{3}} \left((\tilde{x}_{i,j}=N_{\max})\wedge(c_{i,j}^{\min}<\frac{c_{i,j}^{\max}}{3})\right) \right\}
	\end{align}	
\else	
	\begin{align}	
	\label{eq:OmegaD}
		\Omega_{d_1} =& \left\{ (i,j) \left| \vphantom{c_{i,j}^{\min}} \right.\right. \left. c_{i,j}^{\min}+c_{i,j}^{\max}=w^2 \right\}\\
		\Omega_{d_2} =& \left\{ (i,j) \left| \left((\tilde{x}_{i,j}=N_{\min})\wedge(c_{i,j}^{\max}<\frac{c_{i,j}^{\min}}{3})\right) \vee  \right.\right. \left. \vphantom{\frac{c_{i,j}^{\min}}{3}} \left((\tilde{x}_{i,j}=N_{\max})\wedge(c_{i,j}^{\min}<\frac{c_{i,j}^{\max}}{3})\right) \right\}
	\end{align}	
\fi
	where the symbol $\wedge$ denotes logical AND. 
	
	\item 	Compute the set of corrupted pixels $\Omega$ as:
		\begin{align}	
			\Omega = \Omega_N \cap \left( \overline{{\Omega_{d_1}}} \cup \overline{{\Omega_{d_2}}} \right)
			\label{eq:corrupteds}
		\end{align}
		where $\cap$ and $\cup$ denotes the intersection and union operations, respectively, and $\overline{{A}}$ stands for the complement set of $A$. 		
\end{enumerate}

Equations (\ref{eq:OmegaD}-\ref{eq:corrupteds}) imply that a pixel with an impulse value is considered to be uncorrupted if:
\begin{itemize}
	\item All of its neighbors have values equal to the impulse values, 
	\item The majority of its neighbors is inclined to one of the impulse values, 
	\item It contributes in this inclination.
\end{itemize}

We define the mask matrix, corresponding to the set of corrupted pixels $\Omega$, as:
\begin{align}
		\mbox{Mask}_{i,j} = \left\{ \begin{array}{lr} 
														0  & \mbox{if } (i,j)\in\Omega \\
														1  & \mbox{if } (i,j)\notin\Omega
												 \end{array}
											  \right.
		\label{eq:mask}
\end{align}

After impulse detection, we rename the uncorrupted and corrupted pixels as known and noisy pixels, respectively.

\subsection{Image Restoration}
For image restoration, we propose an Efficient Weighted-Average (EWA) filter. In the proposed method, first we construct an initial image using the Nearest Neighboring Interpolation (NNI). In this image, each noisy pixel takes the grey-value of its nearest known pixel. We then improve the initial image by employing a weighted-average filter, which applies different procedures for weighting the known and noisy pixels.

Note that in this subsection, all multiplications and divisions are pixel-wise. 

\subsubsection{Weight Assignment Procedures}
The weight assignment procedures are described in this following.

\paragraph{Weights of Known Pixels}
The weight assignment to known pixels is based on the fact that due to the spatial correlation of image pixels, the information of adjacent pixels overlaps. In other words, two separate pixels have more information than two adjacent pixels. This confirms the observations that, with a fixed noise density, random losses can be restored better than block/burst losses \cite{paper:Ferreira:2001}.

Therefore, to quantify the value of unique information in each known pixel, we should determine the solitariness of that pixel. For this, we define the Information Matrix (IM) as follows:
\begin{align}
\mbox{IM}=\frac{1}{\mbox{Mask}*h}
\label{eq:im}
\end{align}
where * denotes the convolution operation, $h$ is the convolution kernel, which is an all one matrix of size $3\times 3$, and \mbox{Mask} is obtained from (\ref{eq:mask}).

The denominator of (\ref{eq:im}) represents a matrix, which elements specify the number of known pixels in the neighborhood of the corresponding image pixel. The set of neighboring pixels is determined by the kernel $h$. The center pixel is also included in the set of neighboring pixels to avoid infinite information value for a solitary known pixel.

Finally, the weights of known pixels are computed as:
\begin{align}
W^{\mbox{known}}=9\times \mbox{IM}
\end{align}

The weights are normalized such that the lowest weight for known pixels is one. Therefore, the range of $W^{\mbox{known}}$ is between 1 and 9. Note that these values are not valid in the positions of noisy pixels.

\paragraph{Weights of Noisy Pixels}
The weight assignment to noisy pixels is based on this image property that farther image pixels have lesser correlation with each other. Thus in the initial image, noisy pixels which are farther from their nearest known pixel take less accurate value.

We compute the Distance Transform of the image to obtain the Distance Matrix (DM) and the Closest-Pixel Map (CPM). Each element of DM and CPM contain the Euclidean distance of the corresponding image pixel with the nearest known pixel and the index of that pixel, respectively.

The weights of noisy pixels are computed as:
\begin{align}
W^{\mbox{noisy}} = \frac{1}{1+\mbox{DM}}
\label{eq:W-noisy}
\end{align}

Equation (\ref{eq:W-noisy}) implies that $W^{\mbox{noisy}}$ is inversely proportional to the DM and is set to be in the range of $\left(0,\frac{1}{2}\right]$.

\subsubsection{Implementation}
One key factor for fast restoration is matrix implementation. The proposed filter has the advantage that all steps are implemented using matrix-based operations. 
The implementation details of the restoration stage are described in the following: 
\begin{enumerate}
	\item Compute the matrices DM and CPM from the corrupted image $\tilde{x}$. For computing the Euclidean distance transform, we used the fast algorithm described in \cite{paper:Maurer:2003}.
	\item Using the CPM, employ NNI on the image $\tilde{x}$ to obtain the initial image $x^{\mbox{init}}$. 
	\item Construct the overall Weight Matrix (WM) as:
		\begin{align}
      \mbox{WM} = \mbox{Mask} . W^{\mbox{known}} + (1-\mbox{Mask}) . W^{\mbox{noisy}}
		\end{align}
	\item Restore the image as:
			\begin{align}
				y = \frac{\left(x^{\mbox{init}}.\mbox{WM}\right)*h}{\mbox{WM}*h}
			\end{align}
\end{enumerate}

\begin{figure}[!t]%
	\centering
	\includegraphics[width=\figwidth]{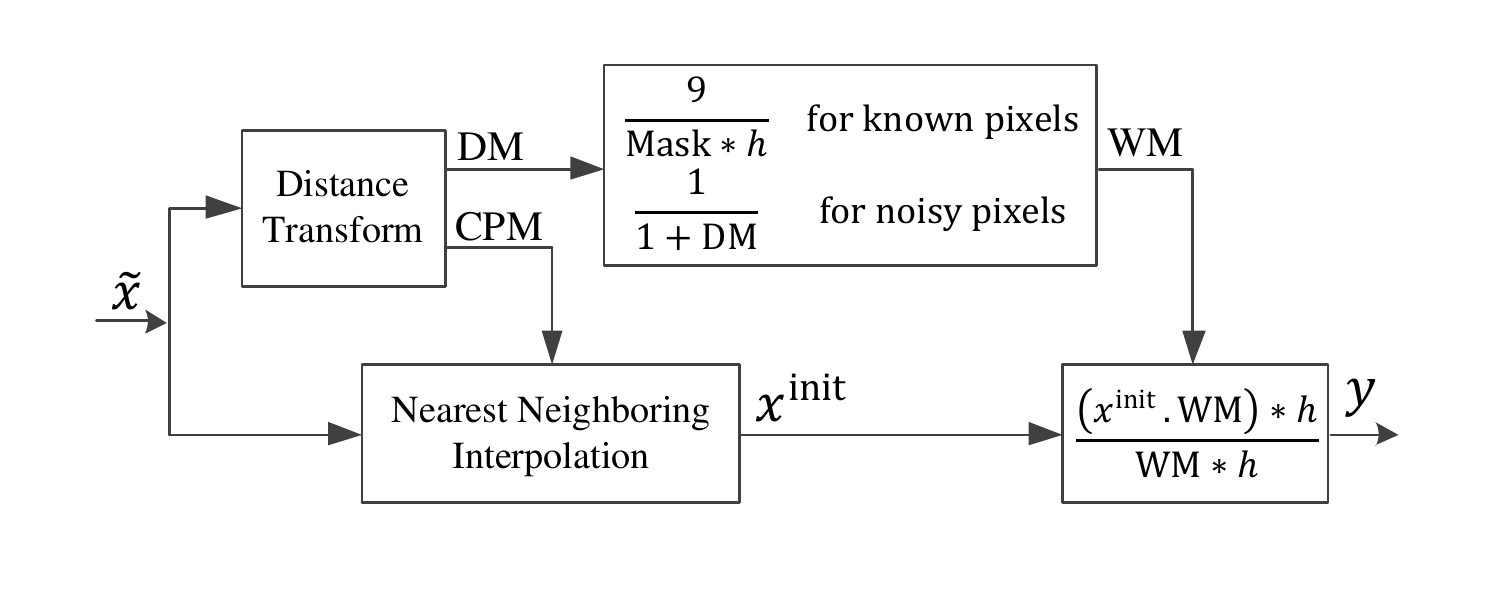}
	\caption{The proposed filter of the restoration stage}%
	\label{fig:impl}%
\end{figure}

Figure \ref{fig:impl} depicts the proposed filter of the restoration stage.

\subsection{Computational Complexity}
In this subsection, we compute the complexity of restoring an $N\times N$ image from r\% SPN, using the EWA filter. 
The major complexity of the proposed filter is for processing NNI, three convolutions, three pixel-wise multiplications and three pixel-wise divisions. The complexity of NNI is $O(N^2)$ \cite{paper:Maurer:2003}. Also, we implement the convolution by two-dimensional FFT (FFT2) using the following relation:
\begin{align}
  a * b = \mbox{FFT}^{-1}\left\{ \mbox{FFT}(a) . \mbox{FFT}(b)  \right\}
\end{align}

The complexity of FFT2 is $2N^2\log⁡N$. As a result, the complexity of computing the two-dimensional convolution of an image with a $3\times 3$ kernel matrix is approximately $4N^2\log⁡N$. Finally, division and multiplication have the complexity of $N^2$. Therefore, regardless of the noise density, the overall complexity is $O(N^2\log⁡N)$. 

\section{Simulation Results}
The proposed algorithm is compared with the best existing methods for SPN removal. Comparisons include the quantitative evaluation, the visual quality and the time complexity. Simulations are run on four $512\times 512$ grey-scale test images {\em Lena, Peppers, Boat} and {\em Bridge}. The Peak Signal-to-Noise Ratio (PSNR) is employed for objective performance assessment. To make a reliable comparison, each method is run 20 times with different impulse noise patterns and the result is obtained by averaging over all experiments.

\ifStyleFinal
	\begin{figure*}[!t]%
		\centering	
		\begin{subfigure}[b]{0.38\textwidth}
			\includegraphics[width=\figwidth]{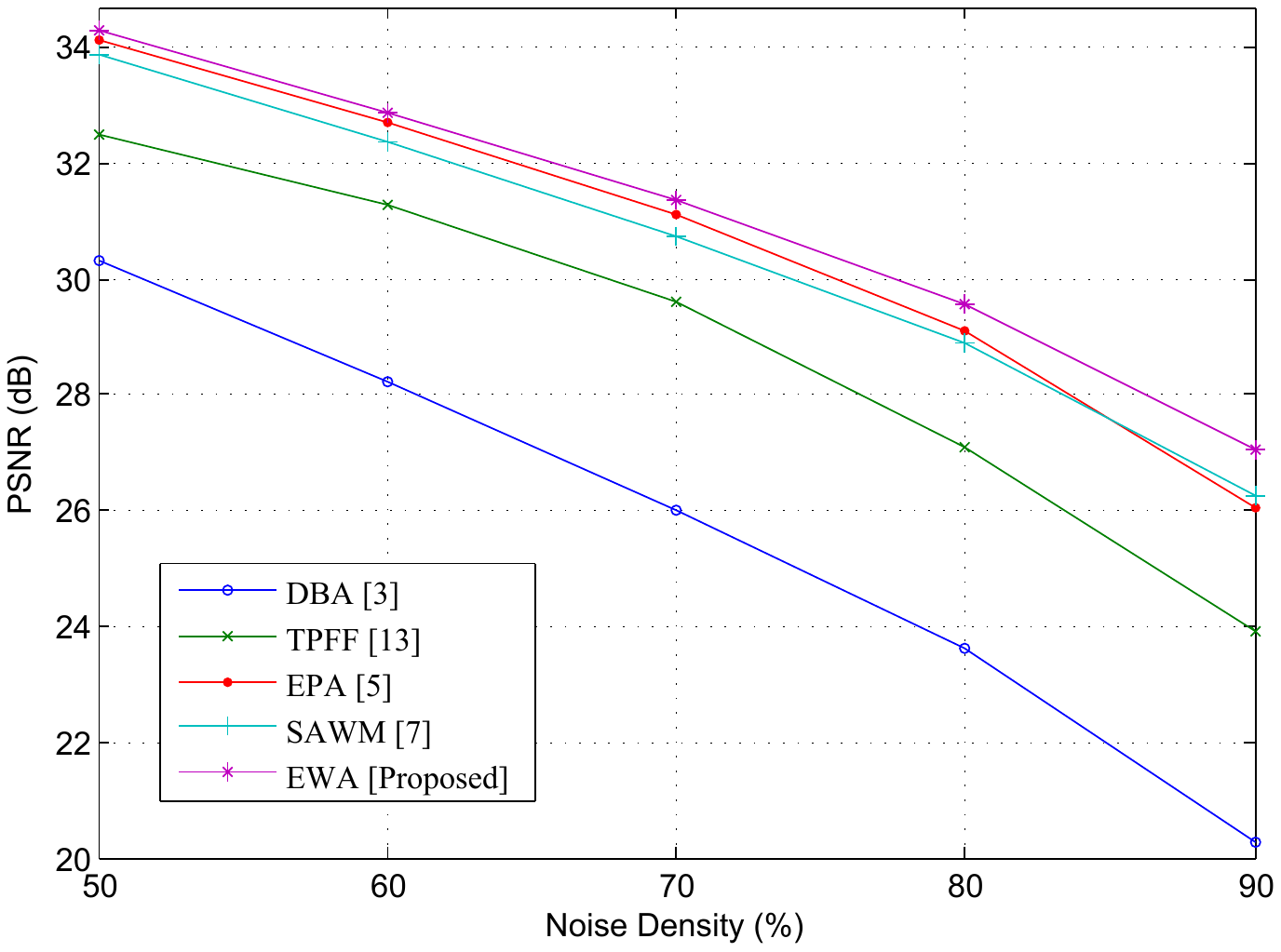}
			\caption{{\em Lena}}
		\end{subfigure}	
		\begin{subfigure}[b]{0.38\textwidth}
			\includegraphics[width=\figwidth]{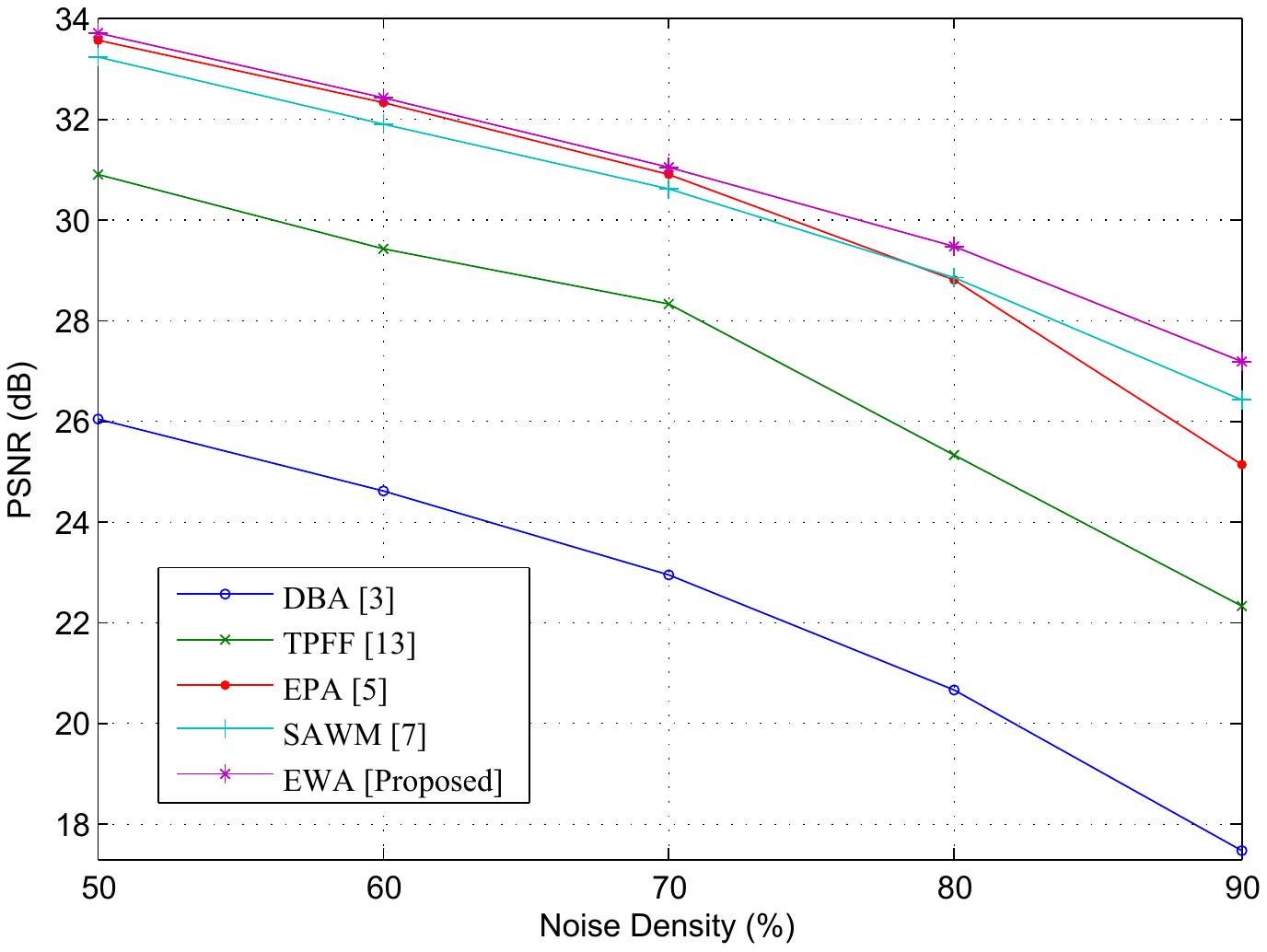}
			\caption{{\em Peppers}}
		\end{subfigure}	
		\begin{subfigure}[b]{0.38\textwidth}
			\includegraphics[width=\figwidth]{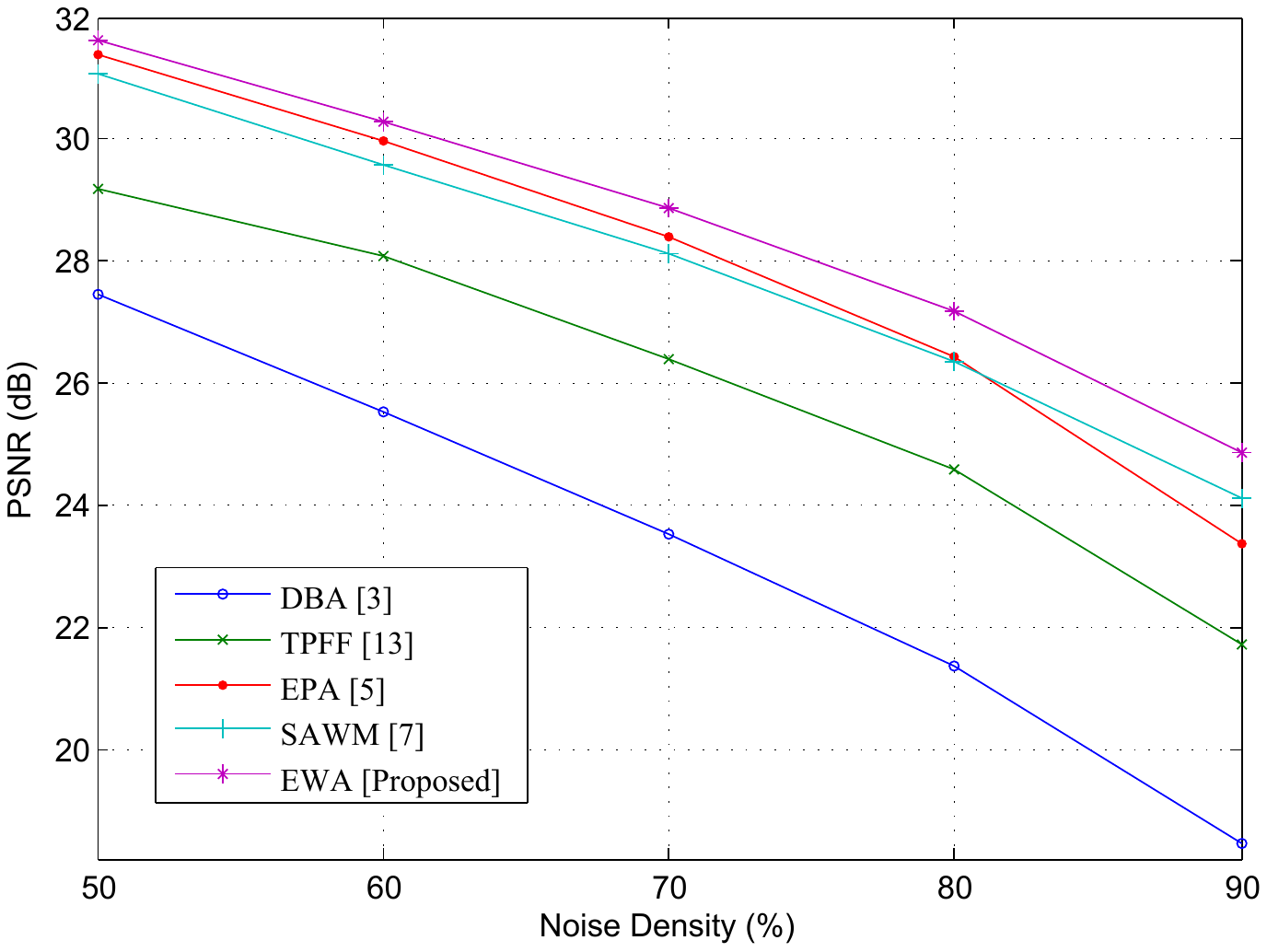}
			\caption{{\em Boat}}
		\end{subfigure}	
		\begin{subfigure}[b]{0.38\textwidth}
			\includegraphics[width=\figwidth]{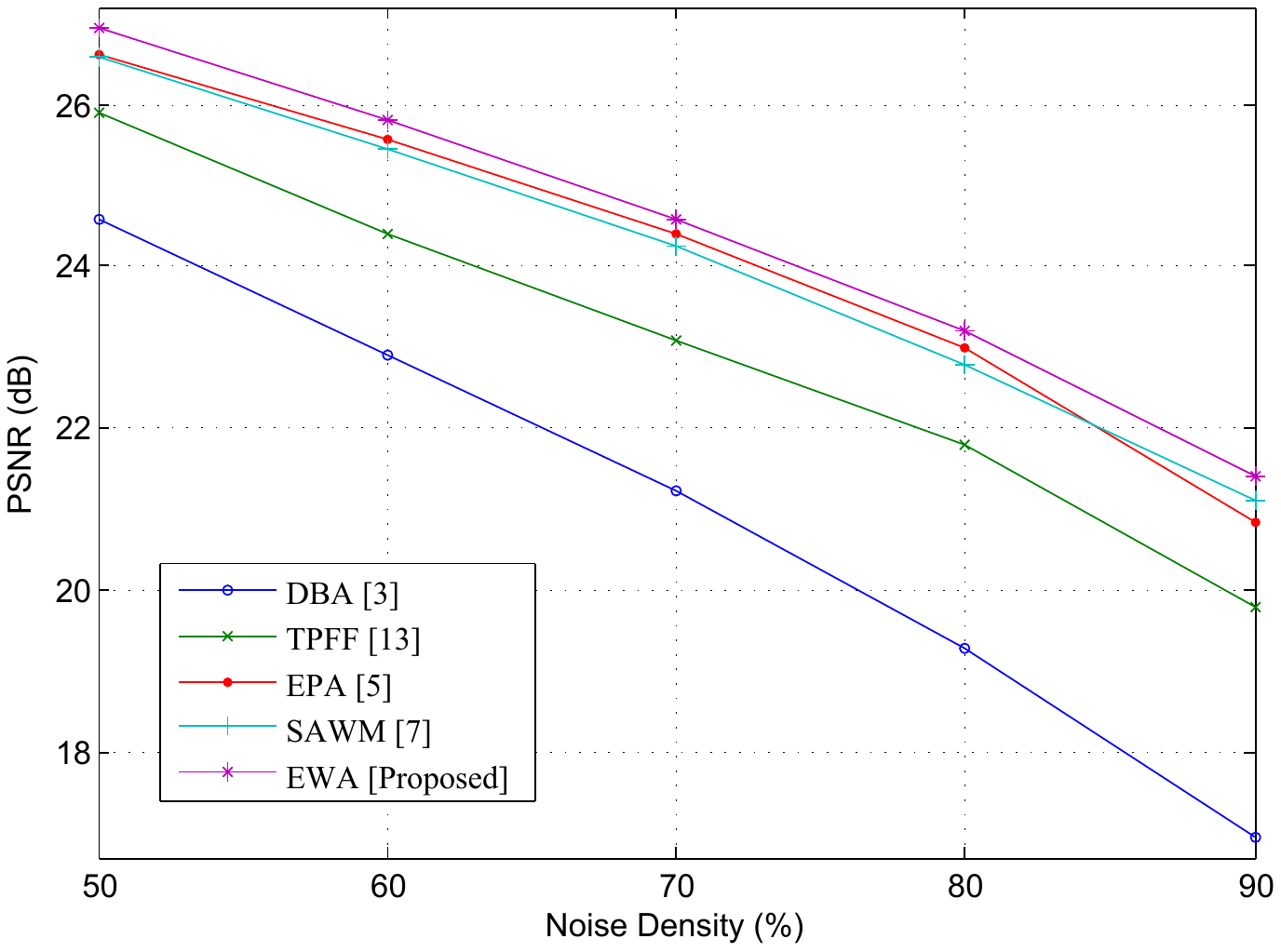}
			\caption{{\em Bridge}}
		\end{subfigure}	
		\caption{Comparison of restoration results in PSNR (dB) for different images corrupted by various densities of SPN.}%
		\label{fig:cmpr}%
	\end{figure*}
\else
	\begin{figure}[H]%
		\centering	
		\subfloat[][{\em Lena}]   {\includegraphics[width=3.5in]{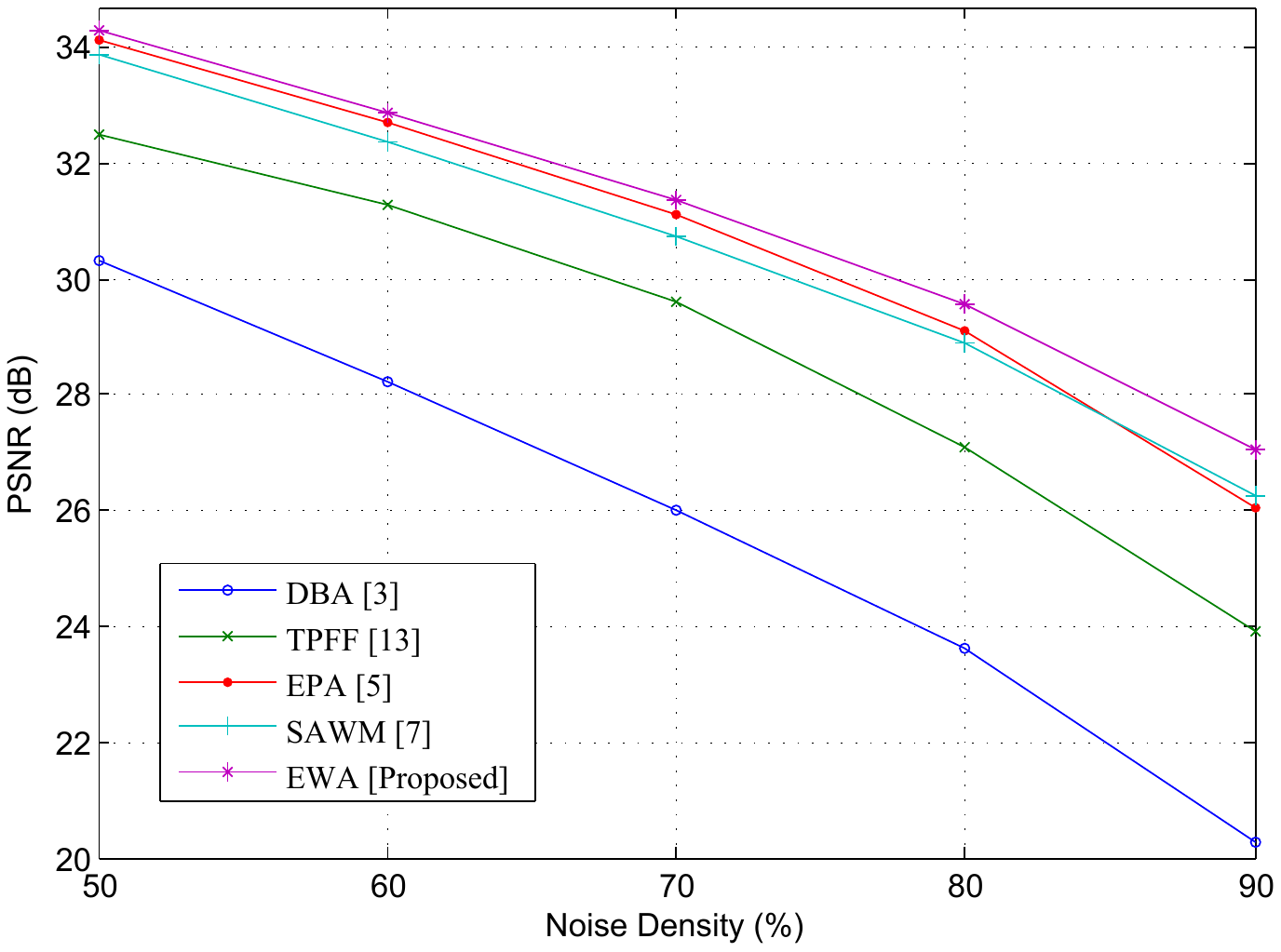}}
		\subfloat[][{\em Peppers}]{\includegraphics[width=3.5in]{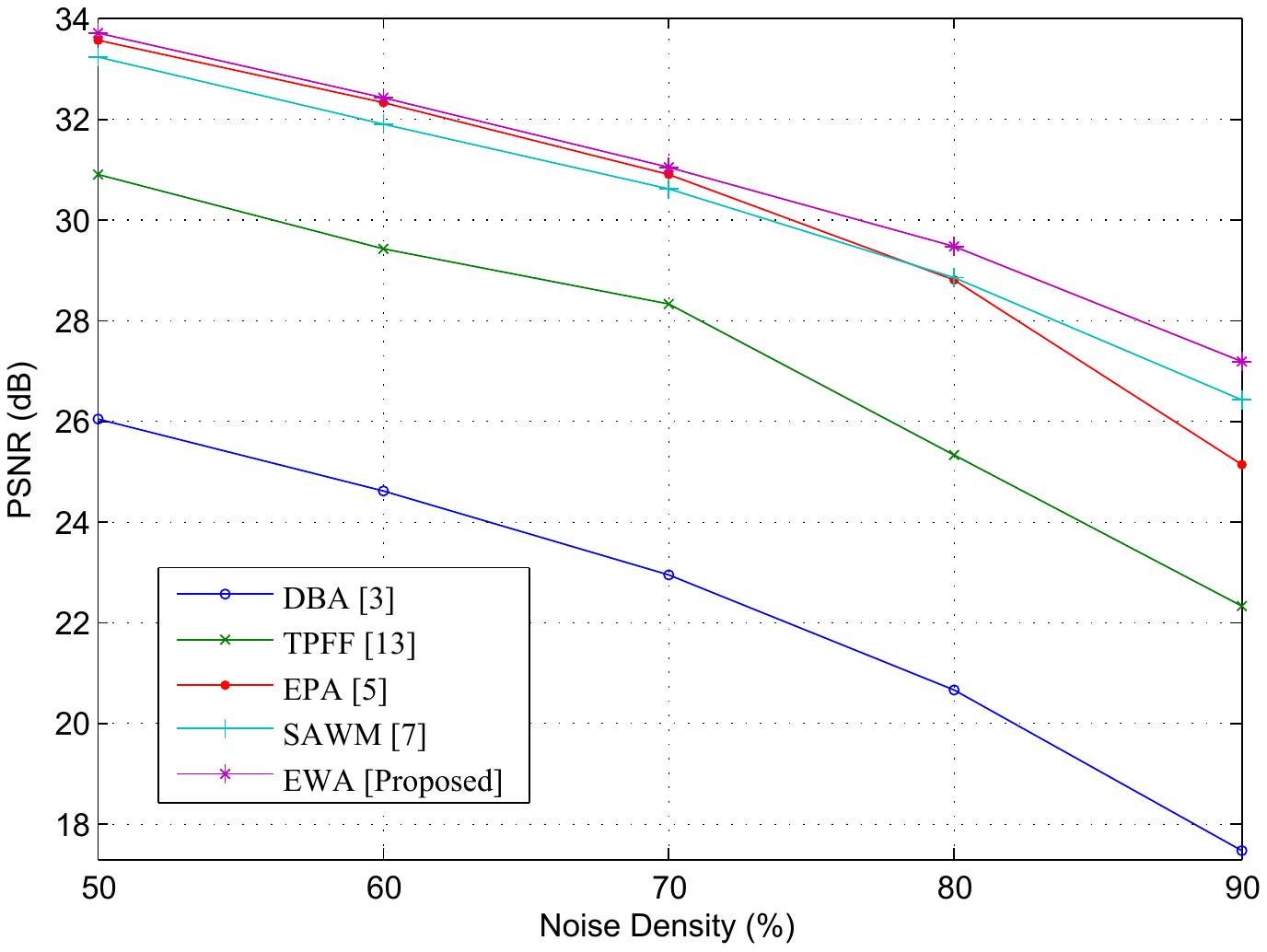}} \\
		\subfloat[][{\em Boat}]   {\includegraphics[width=3.5in]{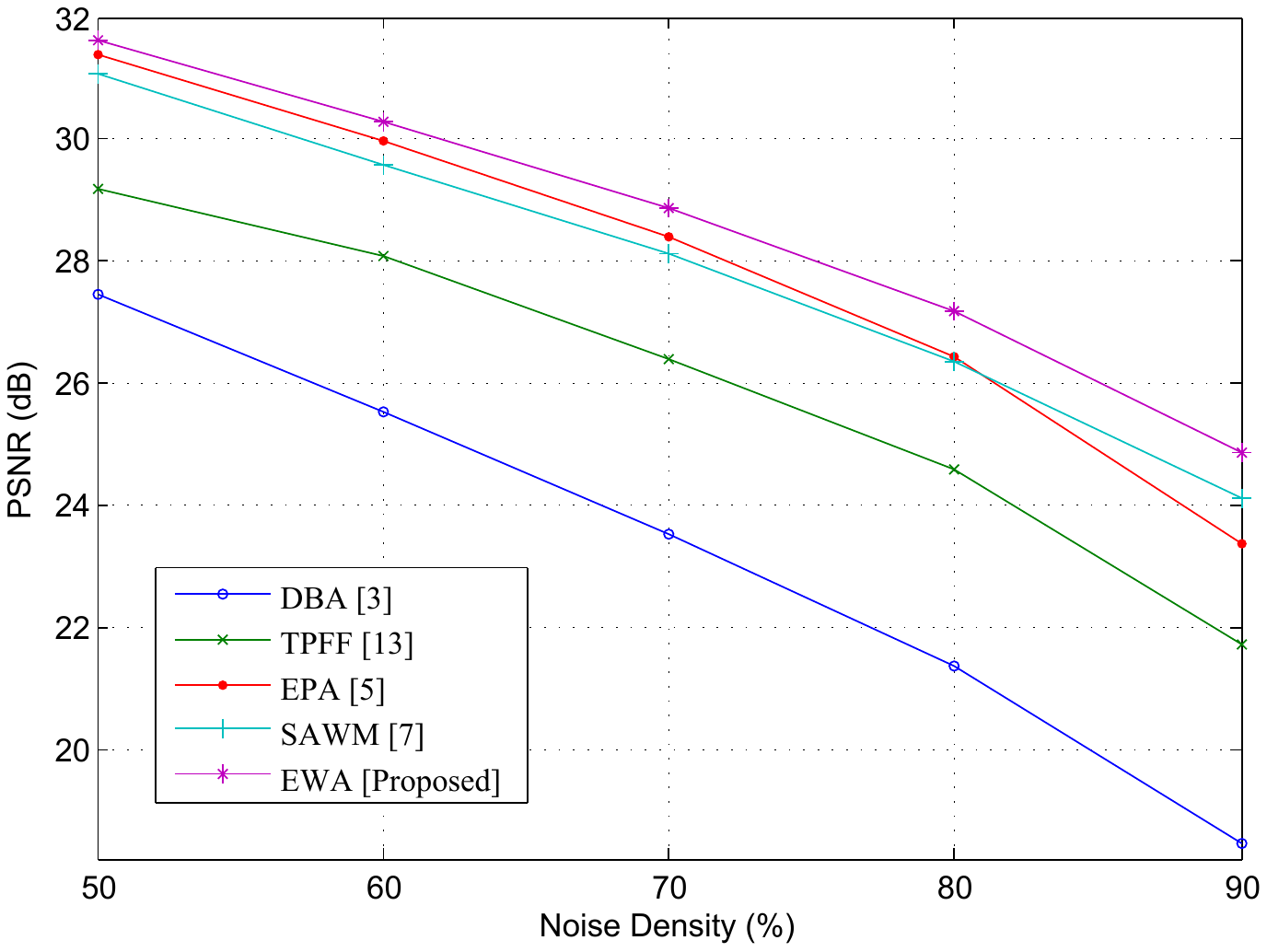}} 
		\subfloat[][{\em Bridge}] {\includegraphics[width=3.5in]{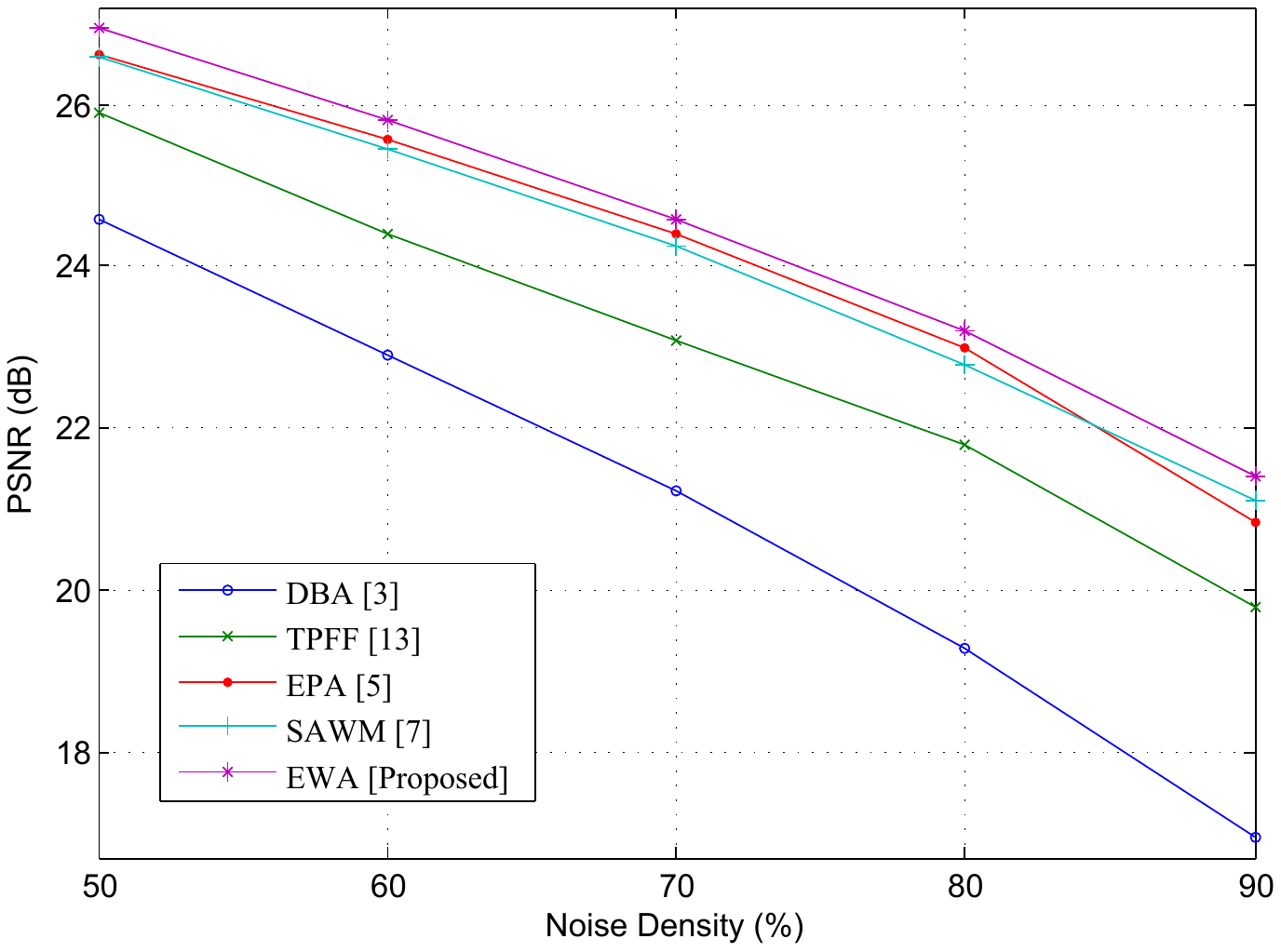}}
		\caption{Comparison of restoration results in PSNR (dB) for different images corrupted by various densities of SPN.}%
		\label{fig:cmpr}%
	\end{figure}
\fi

In section I, we divided the SPN removal methods into four categories. For the sake of brevity, in simulations, we compare the proposed method with the best method of each category. For median-based filters, fuzzy-based algorithms, ad-hoc ideas and weighted-average filters, DBA \cite{paper:Srinivasan:2007}, TPFF \cite{paper:Chou:2013}, EPA \cite{paper:Chen:2008} and SAWM \cite{paper:Zhang:2009} have the best results, respectively. 

Figure \ref{fig:cmpr} depicts the restoration results of different methods for different images corrupted by SPN with various noise densities. The results demonstrate that the EWA filter performs better than other methods. Figures {\ref{fig:lena}-\ref{fig:bridge} exhibit the restored images of the three best filters for images {\em Lena} and {\em Bridge} corrupted by 80\% and 90\% SPN, respectively. Clearly, the proposed method outperforms other methods in preserving the image details. Also, Table \ref{tb:cmpr} lists the running time of the most competitive filters in the MATLAB environment. The simulation results verify that the proposed filter is very efficient for high density impulse noise removal. 

\ifStyleFinal
	\begin{figure*}[!t]%
		\centering	
		\begin{subfigure}[b]{0.23\textwidth}
			\includegraphics[width=\figwidth]{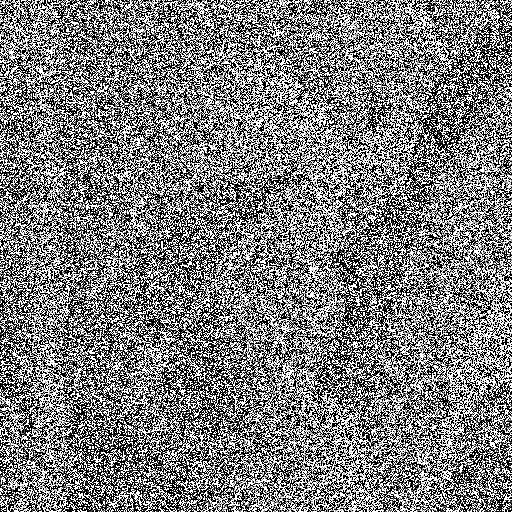}
			\caption{Corrupted image (6.42 dB)}
		\end{subfigure}	
		\begin{subfigure}[b]{0.23\textwidth}
			\includegraphics[width=\figwidth]{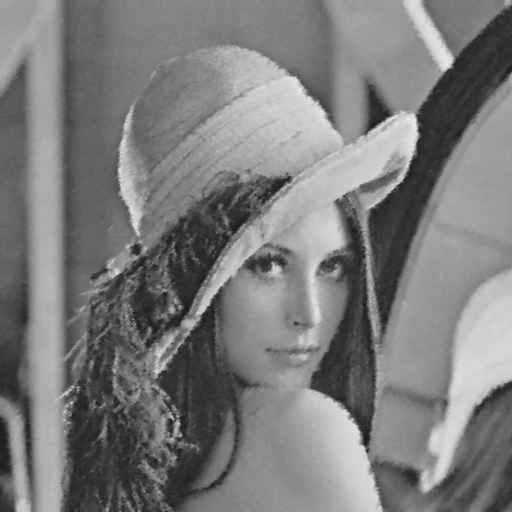}
			\caption{EPA \cite{paper:Chen:2008} (29.11 dB)}
		\end{subfigure}	
		\begin{subfigure}[b]{0.23\textwidth}
			\includegraphics[width=\figwidth]{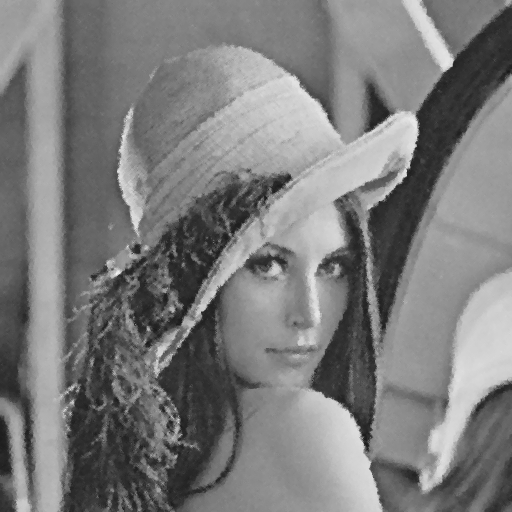}
			\caption{SAWM \cite{paper:Zhang:2009} (28.87 dB)}
		\end{subfigure}	
		\begin{subfigure}[b]{0.23\textwidth}
			\includegraphics[width=\figwidth]{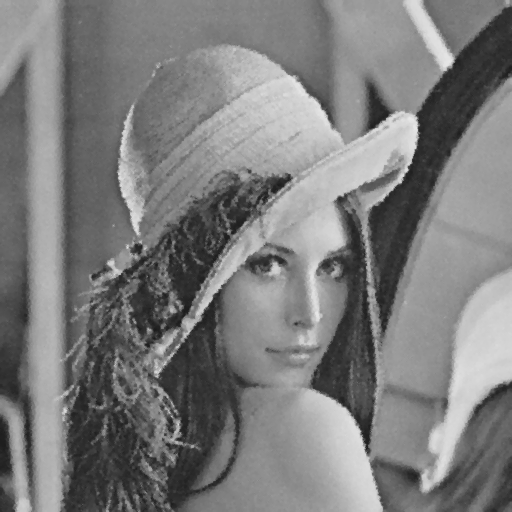}
			\caption{EWA [Proposed] (29.61 dB)}
		\end{subfigure}	
		\caption{Restored images using different filters for image {\em Lena} corrupted by 80\% SPN.}
		\label{fig:lena}%
	\end{figure*}
\else
	\begin{figure*}[H]
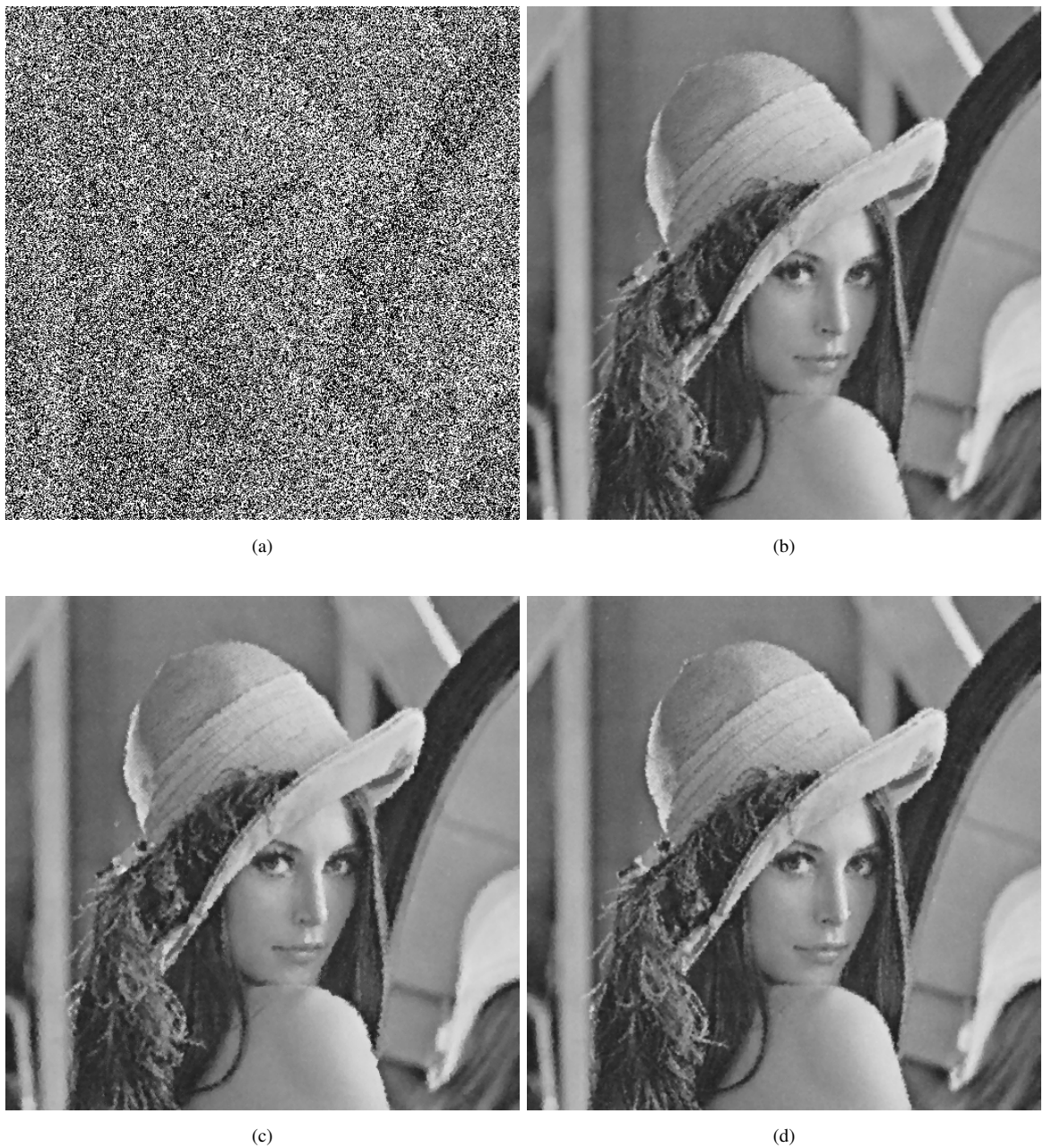
%
		\centering
		\subfloat[][{}]{\includegraphics[width=3.0in]{Fig3-a.png}} \vspace{5pt}
		\subfloat[][{}]{\includegraphics[width=3.0in]{Fig3-b.png}} \\
		\subfloat[][{}]{\includegraphics[width=3.0in]{Fig3-c.png}} \vspace{5pt}
		\subfloat[][{}]{\includegraphics[width=3.0in]{Fig3-d.png}}
		\caption{Restored images using different filters for image {\em Lena} corrupted by 80\% SPN. (a) Corrupted image (6.42 dB), (b) EPA \cite{paper:Chen:2008} (29.11 dB), (c) SAWM \cite{paper:Zhang:2009} (28.87 dB), (d) EWA [Proposed] (29.61 dB)}
		\label{fig:lena}%
	\end{figure*}
\fi

\ifStyleFinal		
	 \begin{figure*}[!t]%
		\centering	
		\begin{subfigure}[b]{0.23\textwidth}
			\includegraphics[width=\figwidth]{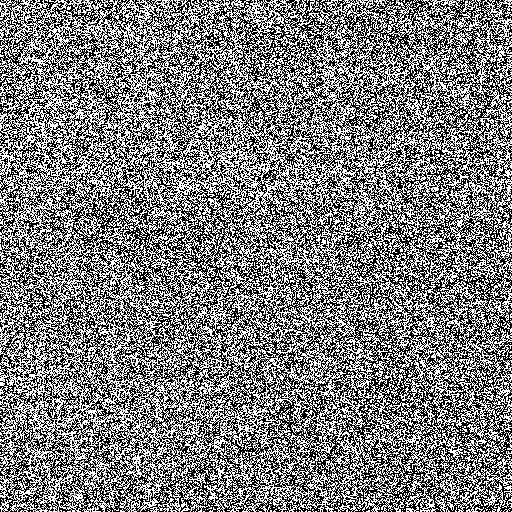}
			\caption{Corrupted image (5.69 dB)}
		\end{subfigure}	
		\begin{subfigure}[b]{0.23\textwidth}
			\includegraphics[width=\figwidth]{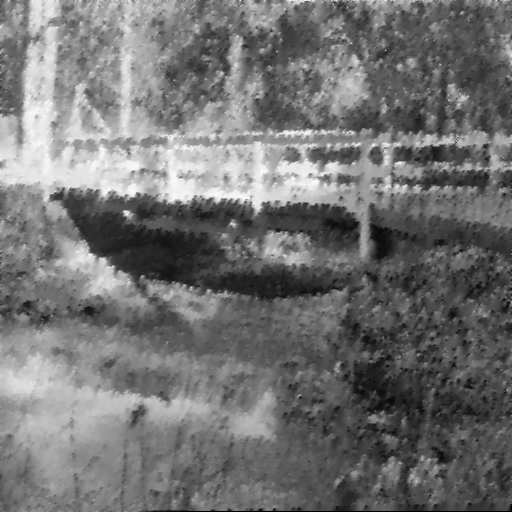}
			\caption{EPA \cite{paper:Chen:2008} (20.76 dB)}
		\end{subfigure}	
		\begin{subfigure}[b]{0.23\textwidth}
			\includegraphics[width=\figwidth]{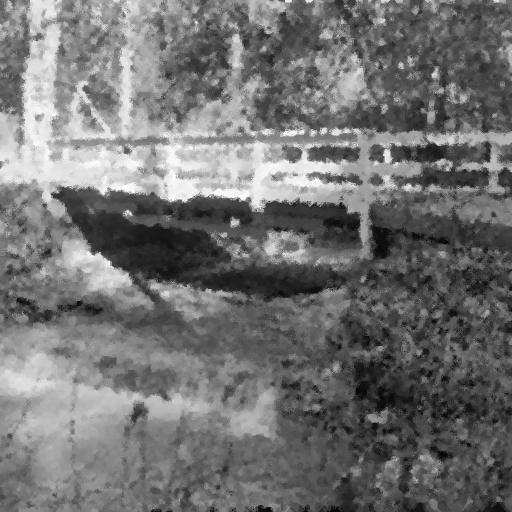}
			\caption{SAWM \cite{paper:Zhang:2009} (21.15 dB)}
		\end{subfigure}	
		\begin{subfigure}[b]{0.23\textwidth}
			\includegraphics[width=\figwidth]{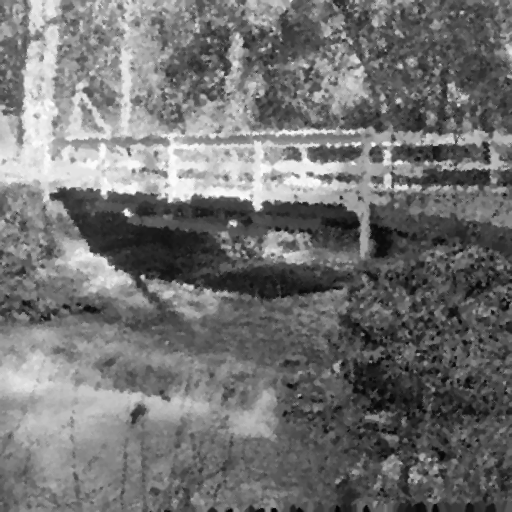}
			\caption{EWA [Proposed] (21.43 dB)}
		\end{subfigure}
		\caption{Restored images using different filters for image {\em Bridge} corrupted by 90\% SPN.}
		\label{fig:bridge}
	\end{figure*}
\else
	\begin{figure*}[H]
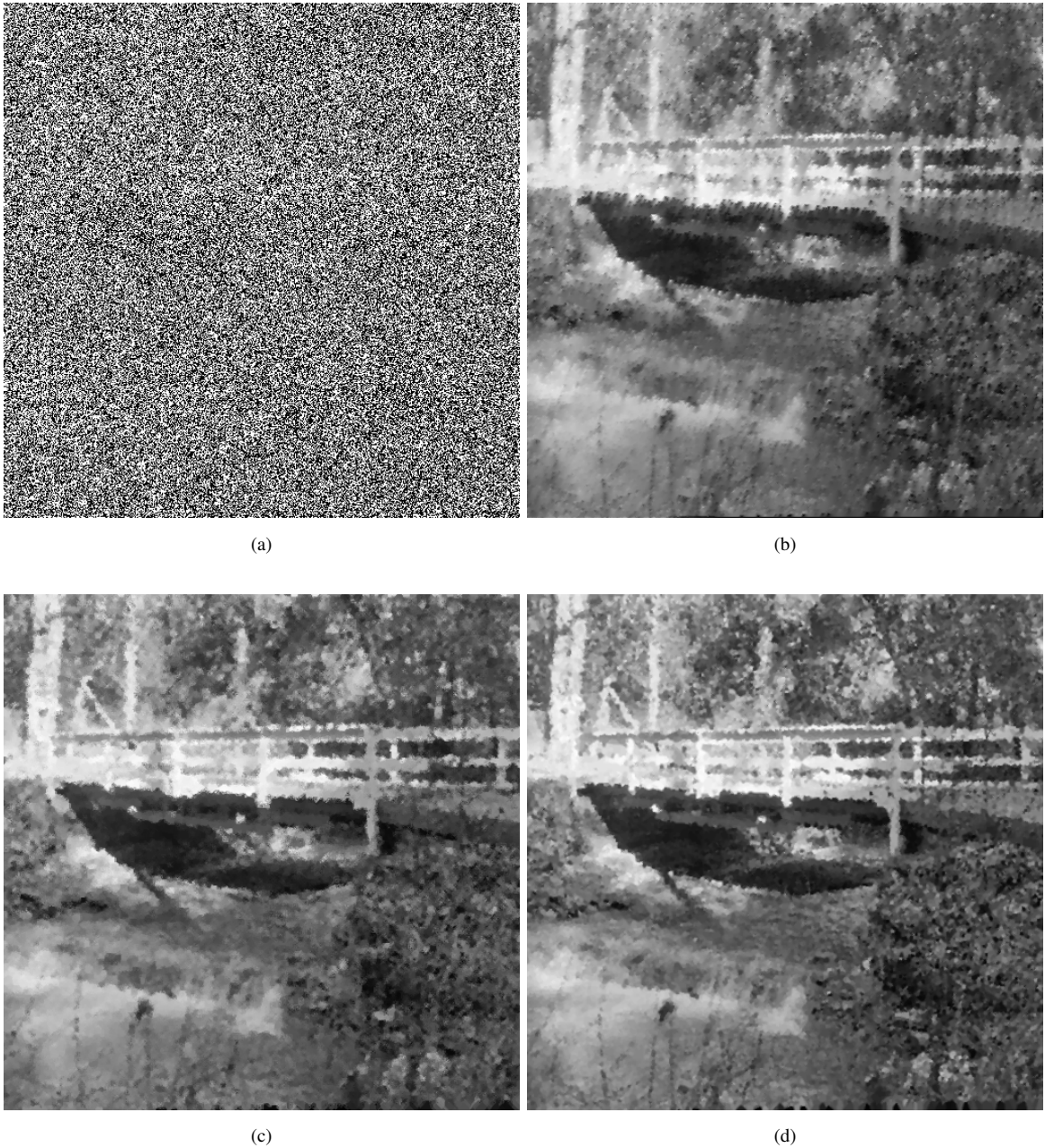
%
		\centering	
		\subfloat[][{}]{\includegraphics[width=3.0in]{Fig4-a.png}} \vspace{5pt}
		\subfloat[][{}]{\includegraphics[width=3.0in]{Fig4-b.png}} \\
		\subfloat[][{}]{\includegraphics[width=3.0in]{Fig4-c.png}} \vspace{5pt}
		\subfloat[][{}]{\includegraphics[width=3.0in]{Fig4-d.png}}
		\caption{Restored images using different filters for image {\em Bridge} corrupted by 90\% SPN. (a) Corrupted image (5.69 dB), (b) EPA \cite{paper:Chen:2008} (20.76 dB), (c) SAWM \cite{paper:Zhang:2009} (21.15 dB), (d) EWA [Proposed] (21.43 dB).}
		\label{fig:bridge}
	\end{figure*}
\fi

\begin{table}%
\centering
\caption{Runtime in Seconds of Different Methods with Various Noise Densities in the MATLAB Environment}
\newcolumntype{Y}{>{\centering\arraybackslash}X}
\begin{tabularx}{\figwidth}{|c|Y|Y|Y|Y|Y|} 
\toprule
\hline 
											& \bf{50\%} & \bf{60\%}	& \bf{70\%}	& \bf{80\%}	& \bf{90\%} \\ [1ex]
\hline
\bf{DBA \cite{paper:Srinivasan:2007}}					& 1.2				& 1.4				& 1.6				& 1.8				& 2.1  \\ [1ex]
\hline 
\bf{EPA \cite{paper:Chen:2008}}					& 0.54			& 0.62			& 0.70			& 0.74			& 0.72 \\ [1ex]
\hline 
\bf{SAWM \cite{paper:Zhang:2009}}					& 10	  		& 11			& 13			& 15			& 16 \\ [1ex]
\hline 
\bf{CM \cite{paper:Zhou:2012}}					& 15			& 22			& 25			& 29			& 39 \\ [1ex]
\hline 
\bf{AIM \cite{paper:Hosseini:2013}}					& 0.14			& 0.16			& 0.19			& 0.25			& 0.31 \\ [1ex]
\hline 
\bf{EWA [Proposed]}		& \bf{0.075} &	\bf{0.075}	&\bf{0.075}	&\bf{0.075}	&\bf{0.075} \\ [1ex]
\hline
\bottomrule
\end{tabularx}
\label{tb:cmpr}
\end{table}

\section{Conclusion}
In this paper, we proposed a method for fast impulse noise removal from images. The proposed filter first constructs an initial image using the nearest neighboring interpolation and then improves it by employing a weighted-average filter, which applies different procedures for weighting the known and noisy pixels. Experimental results verify that the proposed method outperforms the best existing methods in both qualitative and quantitative measures and is quite suitable for real-time applications.

\ifCLASSOPTIONcaptionsoff
  \newpage
\fi

\bibliographystyle{IEEEtran}
\bibliography{IEEEabrv,mylit}

\begin{thebibliography}{10}
\providecommand{\url}[1]{#1}
\csname url@samestyle\endcsname
\providecommand{\newblock}{\relax}
\providecommand{\bibinfo}[2]{#2}
\providecommand{\BIBentrySTDinterwordspacing}{\spaceskip=0pt\relax}
\providecommand{\BIBentryALTinterwordstretchfactor}{4}
\providecommand{\BIBentryALTinterwordspacing}{\spaceskip=\fontdimen2\font plus
\BIBentryALTinterwordstretchfactor\fontdimen3\font minus
  \fontdimen4\font\relax}
\providecommand{\BIBforeignlanguage}[2]{{%
\expandafter\ifx\csname l@#1\endcsname\relax
\typeout{** WARNING: IEEEtran.bst: No hyphenation pattern has been}%
\typeout{** loaded for the language `#1'. Using the pattern for}%
\typeout{** the default language instead.}%
\else
\language=\csname l@#1\endcsname
\fi
#2}}
\providecommand{\BIBdecl}{\relax}
\BIBdecl

\bibitem{book:Bovic:2000}
{A. Bovik}, \emph{{Handbook of Image and Video Processing}}.\hskip 1em plus
  0.5em minus 0.4em\relax {Academic Press}, {2000}.

\bibitem{paper:Chan:2005}
R.~Chan, C.-W. Ho, and M.~Nikolova, ``Salt-and-pepper noise removal by
  median-type noise detectors and detail-preserving regularization,''
  \emph{Image Processing, IEEE Transactions on}, vol.~14, no.~10, pp.
  1479--1485, Oct 2005.

\bibitem{paper:Srinivasan:2007}
K.~S. Srinivasan and D.~Ebenezer, ``A new fast and efficient decision-based
  algorithm for removal of high-density impulse noises,'' \emph{Signal
  Processing Letters, IEEE}, vol.~14, no.~3, pp. 189--192, March 2007.

\bibitem{paper:Deng:2007}
D.~Ze-Feng, Y.~Zhou-ping, and X.~You-lun, ``High probability impulse
  noise-removing algorithm based on mathematical morphology,'' \emph{Signal
  Processing Letters, IEEE}, vol.~14, no.~1, pp. 31--34, Jan 2007.

\bibitem{paper:Chen:2008}
P.-Y. Chen and C.-Y. Lien, ``An efficient edge-preserving algorithm for removal
  of salt-and-pepper noise,'' \emph{Signal Processing Letters, IEEE}, vol.~15,
  pp. 833--836, 2008.

\bibitem{paper:Yildirim:2008}
M.~Yildirim, A.~Basturk, and M.~Yuksel, ``Impulse noise removal from digital
  images by a detail-preserving filter based on type-2 fuzzy logic,''
  \emph{Fuzzy Systems, IEEE Transactions on}, vol.~16, no.~4, pp. 920--928, Aug
  2008.

\bibitem{paper:Zhang:2009}
X.~Zhang and Y.~Xiong, ``Impulse noise removal using directional difference
  based noise detector and adaptive weighted mean filter,'' \emph{Signal
  Processing Letters, IEEE}, vol.~16, no.~4, pp. 295--298, April 2009.

\bibitem{paper:Toh:2010}
K.~Toh and N.~Isa, ``Noise adaptive fuzzy switching median filter for
  salt-and-pepper noise reduction,'' \emph{Signal Processing Letters, IEEE},
  vol.~17, no.~3, pp. 281--284, March 2010.

\bibitem{paper:Fabijanska:2011}
A.~Fabijańska and D.~Sankowski, ``Noise adaptive switching median-based
  filter for impulse noise removal from extremely corrupted images,''
  \emph{Image Processing, IET}, vol.~5, no.~5, pp. 472--480, August 2011.

\bibitem{paper:Esakkirajan:2011}
S.~Esakkirajan, T.~Veerakumar, A.~Subramanyam, and C.~PremChand, ``Removal of
  high density salt and pepper noise through modified decision based
  unsymmetric trimmed median filter,'' \emph{Signal Processing Letters, IEEE},
  vol.~18, no.~5, pp. 287--290, May 2011.

\bibitem{paper:Xiong:2012}
B.~Xiong and Z.~Yin, ``A universal denoising framework with a new impulse
  detector and nonlocal means,'' \emph{Image Processing, IEEE Transactions on},
  vol.~21, no.~4, pp. 1663--1675, April 2012.

\bibitem{paper:Zhou:2012}
Z.~Zhou, ``Cognition and removal of impulse noise with uncertainty,''
  \emph{Image Processing, IEEE Transactions on}, vol.~21, no.~7, pp.
  3157--3167, July 2012.

\bibitem{paper:Chou:2013}
H.-H. Chou, L.-Y. Hsu, and H.-T. Hu, ``Turbulent-pso-based fuzzy image filter
  with no-reference measures for high-density impulse noise,''
  \emph{Cybernetics, IEEE Transactions on}, vol.~43, no.~1, pp. 296--307, Feb
  2013.

\bibitem{paper:Hosseini:2013}
H.~Hosseini and F.~Marvasti, ``Fast restoration of natural images corrupted by
  high-density impulse noise,'' \emph{EURASIP J. Image and Video Processing},
  2013.

\bibitem{code:Hossein:2014}
\BIBentryALTinterwordspacing
H.~Hosseini. (2014) Ewa implementaion, source code and results. [Online].
  Available: \url{https://github.com/HosseinHosseini/EWA}
\BIBentrySTDinterwordspacing

\bibitem{paper:Ferreira:2001}
\BIBentryALTinterwordspacing
P.~Ferreira, ``\BIBforeignlanguage{English}{Iterative and noniterative recovery
  of missing samples for 1-d band-limited signals},'' in
  \emph{\BIBforeignlanguage{English}{Nonuniform Sampling}}, ser. Information
  Technology: Transmission, Processing, and Storage, F.~Marvasti, Ed.\hskip 1em
  plus 0.5em minus 0.4em\relax Springer US, 2001, pp. 235--281. [Online].
  Available: \url{http://dx.doi.org/10.1007/978-1-4615-1229-5_5}
\BIBentrySTDinterwordspacing

\bibitem{paper:Maurer:2003}
J.~Maurer, C.R., R.~Qi, and V.~Raghavan, ``A linear time algorithm for
  computing exact euclidean distance transforms of binary images in arbitrary
  dimensions,'' \emph{Pattern Analysis and Machine Intelligence, IEEE
  Transactions on}, vol.~25, no.~2, pp. 265--270, Feb 2003.

\end{thebibliography}

\end{document}